\documentclass[pmlr]{jmlr}

\usepackage[load-configurations=version-1]{siunitx} 
\usepackage{longtable}

 \usepackage{booktabs}
\usepackage{pifont}
\newcommand{\cmark}{\ding{51}}%
\newcommand{\xmark}{\ding{55}}%
\usepackage[load-configurations=version-1]{siunitx} 

\makeatletter
\def\set@curr@file#1{\def\@curr@file{#1}} 
\makeatother

\newcommand{\indep}{\rotatebox[origin=c]{90}{$\models$}}

\usepackage{listings}
\usepackage{cleveref}
\usepackage{soul}

\theorembodyfont{\upshape}
\theoremheaderfont{\scshape}
\theorempostheader{:}
\theoremsep{\newline}

\jmlrvolume{182}
\jmlryear{2022}
\jmlrworkshop{Machine Learning for Healthcare}

\title[Survival MDN]{Survival Mixture Density Networks}

\author{\Name{Xintian Han}
       \Email{xintian.han@nyu.edu}\\ 
       \addr New York University
       \AND
       \Name{Mark Goldstein}
       \Email{goldstein@nyu.edu}\\ 
       \addr New York University
        \AND
       \Name{Rajesh Ranganath}
       \Email{rajeshr@cims.nyu.edu}\\ 
       \addr New York University} 

\begin{document}

\maketitle

\begin{abstract}
Survival analysis, the art of time-to-event modeling, plays an important role in clinical treatment decisions. Recently,  continuous time models built from neural ODEs have been proposed for survival analysis. However, the training of neural ODEs is slow due to the high computational complexity of neural ODE solvers. Here, we propose an efficient alternative for flexible continuous time models, called Survival Mixture Density Networks (Survival MDNs). 
Survival MDN applies an invertible positive function to the output of Mixture Density Networks (MDNs). While MDNs produce flexible real-valued distributions, the invertible positive function maps the model into the time-domain while preserving a tractable density. Using four datasets, we show that Survival MDN performs better than, or similarly to
continuous and discrete time
baselines on concordance, integrated Brier score and integrated binomial log-likelihood. Meanwhile, Survival MDNs are also faster than ODE-based models and circumvent binning issues in discrete models.
\end{abstract}

\section{Introduction}
Survival analysis serves as an important tool in healthcare to assess the risk of events, such as onset of disease~\citep{wilson1998prediction} or death~\citep{pocock1982long},  rehospitalization~\citep{patterson1998intensive} and  discharge from hospital~\citep{wang2020survival}. Survival modeling has been widely used in clinical applications, including improving the prognosis of cancer~\citep{faradmal2012survival, goldstraw2016iaslc, wang2019clinicopathological, lin2021effects, wang2021survnet}, predicting the onset of septic shock~\citep{henry2015targeted}, assessing the survival time of heart failure patients~\citep{ahmad2017survival, kojoria2004outcomes, jones2019survival, yin2022survival} and estimating the graft survival rate of kidney transplant patients~\citep{lee2019long, rodrigues2019survival}. 

Given patients' electronic health records including lab tests, vitals, radiology results and clinical notes, doctors need to determine a level of treatment based on the level of risk. For example, WHO guidelines suggest more aggressive treatments for higher risk cardiovascular disease patients~\citep{world2007prevention}. Therefore, an accurate model of risk is necessary. 

Risk in survival analysis is characterized by the conditional distribution of the event time given a patient's healthcare records. What distinguishes survival analysis 
from traditional regression problems is that event times can be censored, i.e., only known to lie within a certain range. For example, patients may remain healthy throughout a 10-year coronary artery disease study~\citep{wilson1998prediction} so it is only known that such patients survive at least 10 years. Discarding censored times may introduce bias into estimates by underestimating the time until an event, because later times are more likely to be censored and thus thrown away.

Likelihood-based methods are used to estimate survival models \citep{kalbfleisch2011statistical}. 
In addition to the usual mass or density computed in maximum likelihood problems, the survival likelihood for censored data includes the survival function, i.e., one minus the cumulative distribution function (CDF) of the distribution. For many distributions, CDF evaluations require explicitly integrating the density.  Recent advances in deep learning provide opportunities for flexible survival modeling~\citep{lecun2015deep, ranganath2016deep}. However,  flexible distributions utilizing deep learning, such as those modeled by GANs~\citep{goodfellow2014generative, chapfuwa2018adversarial}, may not yield efficient CDF computation.

To keep estimation tractable, traditional survival analysis techniques make distributional assumptions, e.g. log-normal density or proportional hazards \citep{kalbfleisch2011statistical, cox1972regression}. But this limits the flexibility of the model. To move beyond this, discrete time models divide continuous times into a sequence of bins~\citep{miscouridou2018deep, lee2018deephit, kvamme2019continuous} and can 
approximate arbitrary continuous distributions increasingly
well as the number of bins increases.
 However, the choice of bin boundaries is troublesome: it is unclear how best to set the time intervals for each bin, 
and the survival function for times within a bin is ill-defined. ODE-based continuous time models~\citep{tang2020soden, tang2022survival} specify the time-to-event distribution through ODEs. However, the training of ODE-based models is slow due to expensive numerical integration requiring many neural network evaluations for each forward pass~\citep{kelly2020learning}.

In this work, we propose Survival Mixture Density Networks (Survival MDN). Survival MDN builds off mixture density networks (MDN) \citep{bishop1994mixture} to allow flexible modeling. Since the time-to-event is positive in survival modeling, we apply an invertible positive function to the samples from MDNs. The CDF of Survival MDNs can be obtained easily through the evaluation of the CDF of
the mixture components of the MDN, which is simple for mixture components like Gaussians. We evaluate Survival MDN and baselines on four clinical datasets: SUPPORT, METABRIC, GBSG, and MIMIC. On all datasets, Survival MDN performs better than, or as well as, the baselines on concordance, integrated Brier Score and integrated binomial log-likelihood. We also show that training Survival MDNs can be 100 times faster than the ODE-based model SODEN~\citep{tang2020soden}.\footnote{The code is available at https://github.com/XintianHan/Survival-MDN}

\subsection*{Generalizable Insights about Machine Learning in the Context of Healthcare}
The majority of flexible survival modeling relies on training with the Cox partial likelihood, discrete time modeling, or ordinary differential equations. Training with partial likelihood
precludes the use of stochastic gradient descent and is not scalable for large datasets. Discrete time models have issues with choosing bin boundaries and determining the survival probability for a particular time. ODE-based models use likelihood for training but are slow to train.
Our proposed model Survival MDN have several advantages 1) It is a continuous time model 2) It makes fewer distributional assumptions
3) It can be trained with stochastic gradient descent 4) It is easier to use than discrete models and faster than ODE-based models.

\section{Background}
In this section, we introduce the mathematical foundation of survival analysis and summarize related works. We then describe how our work is distinguished from previous works.
\subsection{Foundation of Survival Analysis}
Survival analysis studies the distribution of event time $T$ given covariates $X$.
For example, we would like to know when a patient may die after the admission to ICU. The event time is called the failure time or survival time. We consider the common scenario of \textit{right-censoring} in this work, where only a lower bound of the survival time is observed for some patients. We call the lower bound the \textit{censored time} $C$. When $T > C$, only the censored time $C$ is observed; when $T\leq C$,the failure time $T$ is observed. We use $\Delta = I\{T\leq C\}$ to indicate whether the event time is observed and $U = \min\{T, C\}$ to denote the observed time.

A central quantity that appears in the estimation and use of survival models is the survival function $S(t|X) = P (T>t|X)$, i.e., the probability a patient with covariates $X$ will survival until time $t$. By definition, $S(t|X) = 1 - \text{CDF}(t|X)$. 

Assume we observe i.i.d. datapoints $\{u_i, \Delta_i, x_i\}_{i=1}^N$ and censoring is random $T\perp C |X$. Under these assumptions, 
and with $p(t|X)$ denotating the mass or probability density function (PDF) evaluated at $t$,
the survival likelihood function with a parameter $\theta$ is proportional to~\citep{kalbfleisch2011statistical}:
\[
\Pi_{i=1}^N p_\theta(u_i|x_i)^{\Delta_i}S_\theta(u_i|x_i)^{1-\Delta_i}.
\]
In this work, we use the log-likelihood as a training objective function.
\subsection{Related Work}
\paragraph{Traditional Survival Analysis} Traditionally, survival analysis makes distributional assumptions. The Cox model~\citep{cox1972regression} makes the proportional hazard assumption. The accelerated failure time (AFT) model~\citep{buckley1979linear,wei1992accelerated} assumes that $\log(T) = X^T\theta +\epsilon$, where $\epsilon$ follows a log-logistic distribution. Multiple variants of Cox and AFT models~\citep{aalen1980model, bennett1983analysis, cheng1995analysis, lin1995semiparametric, kalbfleisch2011statistical, wu2019flexible} have been proposed to introduce time-varying functions or different distributions. However, these extensions only use linear or simple non-linear models which may not be flexible enough to model complex data distributions. \citet{avati2020countdown} use deep networks
to produce the parameters of a lognormal.
Though this can capture nonlinear dependence of the lognormal's parameters on the input,
the lognormal assumption may not be appropriate, e.g., if the true
conditional distribution has more than one mode.

\paragraph{Deep Cox Models}
The Cox model has been extended with deep networks
in several ways. DeepSurv~\citep{katzman2018deepsurv} uses a neural network to model the relative risk $g(X;\theta)$. Cox-Time~\citep{kvamme2019time} further allows the relative risk to depend on time $t$. \citet{kvamme2019continuous} assume the hazard is constant in predefined time intervals. \citet{nagpal2021deep} uses a mixture of Cox models parameterized by neural networks. These models optimize the partial likelihood function which does not require the access to survival functions. The partial likelihood is defined by
\[
\Pi_{i:\Delta_i = 1}\frac{\exp(g(u_i, x_i;\theta))}{\sum_{j\in R_i} \exp(g(u_i, x_j;\theta))},
\]
where $R_i = \{j: y_j \geq y_i\}$, called the risk set, denotes the set of patients who survive at least as long as the $i$-th patient. The goal of maximizing the partial likelihood is to make patient $i$'s relative risk at $u_i$ greater than that of the other patients who survive longer. 
When there are thousands of datapoints, stochastic gradients are
an efficient alternative to gradient computation for maximizing likelihoods.
However, risk sets require the whole dataset to evaluate since the risk set involves all patients. This disadvantage precludes the use of stochastic gradient descent for training. Though we can use mini-batches of patients to approximate the risk set $R_i$, there are no theoretical guarantees for convergence.

\paragraph{Deep Discrete Models} 
Deep categorical survival models~\citep{miscouridou2018deep,fotso2018deep,goldstein2020x} divide the time axis into a sequence of bins and turn survival analysis into predicting a time's bin. These models use $K$ bins where the last
bin includes all times greater than some value. DeepHit~\citep{lee2018deephit} adds a rank-based loss and uses discrete models for competing risks. Nnet-survival~\citep{biganzoli1998feed,gensheimer2019scalable} models the survival function by multiplications of conditional probabilities in previous time bins.
These discrete models can approximate
arbitrary smooth distributions with increasing fidelity as $K$ increases \citep{miscouridou2018deep}.

However, discrete models have their own problems. These models do not define what happens to the survival function estimation within a bin, at least without additional assumptions e.g. linearly interpolating the CDF. Next, it is challenging to choose the bin boundaries; it is unclear whether to set them by population percentiles or by regular intervals \citep{kvamme2019continuous, tang2020soden, craig2021survival}. 
Using regular intervals may lead times to concentrate into a small subset of bins. For percentiles, it is unclear whether we should include the censored times into the population. Percentiles of the observed failure times may not equal the percentiles of true failure times. Finally, deep discrete models are based on classification architectures, meaning that they may be overconfident and suffer the same poor calibration observed for deep classifiers \citep{guo2017calibration}, as shown for survival analysis in \cite{goldstein2020x}.

\paragraph{ODE-based Models} Recently, continuous time models with neural ODEs have been proposed
\citep{chen2018neural}. SODEN~\citep{tang2020soden} considers the evolution of cumulative hazard functions as an ODE while \citet{danks2022derivative} model the CDF by an ODE. \citet{groha2020general} use ODEs for multi-state survival analysis. ODE-based models have tractable PDFs and CDFs. However, training neural ODEs is slow~\citep{kelly2020learning} because of the expensive numerical integration inside ODE solvers. ODE-based models also involve extra hyperparameters related to ODE-solvers, including the solver type and tolerance level.

\paragraph{Other Deep Models} \citet{chapfuwa2018adversarial}
use GANs for survival distribution modeling. But they do not use the likelihood as an objective for training since the PDF and CDF of GANs are intractable. The alternative, minimax training of GANs, is known to be unstable~\citep{kodali2017convergence, bottou2018geometrical}. \citet{ranganath2016deep} use deep exponential families~\citep{ranganath2015deep} with Weibull likelihoods.
This approach necessitates the use of black-box variational inference with Monte Carlo gradients \citep{ranganath2014black, mohamed2020monte}, which typically yields both a lower bound on the likelihood and noisier, slower optimizations. Survival stacking~\citep{craig2021survival} casts the survival analysis as a classification task by predicting whether one patient is in other patients' risk sets. But for $N$ datapoints, survival stacking creates $O(N^2)$ classification problems which is not tractable for large datasets.
\paragraph{Our Model} In this work, we propose a new flexible survival model named Survival Mixture Density Networks. Survival MDNs utilize mixture density networks~\citep{bishop1994mixture} to allow flexible modeling. With Gaussians as the base distributions, computing the model CDF and PDF requires the evaluation of standard functions and the error function. The error function can be obtained efficiently via common approximations~\citep{abramowitz1988handbook} and Gaussian CDFs are implemented
in most packages. Our simple approach can be trained through stochastic gradient descent and much faster than ODE-based models. We compare our model with previous approaches in \cref{tab:comp}.
\begin{table}[h]
    \centering
    \begin{tabular}{ccccc}
    \toprule
      Model   &  Flexible & Continuous-time & SGD & Without ODE-Solver\\ \midrule
       Cox  &  \xmark & \cmark & \xmark & \cmark\\
       DeepSurv &  \xmark & \cmark & \xmark & \cmark\\
       DeepHit & \cmark & \xmark & \cmark & \cmark\\
       Nnet-survival & \cmark & \xmark & \cmark & \cmark\\
       Cox-Time & \cmark & \cmark & \xmark & \cmark\\
       SODEN & \cmark & \cmark & \cmark & \xmark\\
       Survival MDN & \cmark & \cmark & \cmark & \cmark \\  \bottomrule
    \end{tabular}
    \caption{Comparison of Different Models}
    \label{tab:comp}
\end{table}
In summary, we propose a continuous-time model that can be trained with stochastic gradients, without numerical ODE solving, and that moves beyond common modeling restrictions (e.g. that the density is log-normal or Cox).

\section{Survival Mixture Density Networks}
Our purpose is to build a survival model that has the following properties:
\begin{enumerate}
    \item It has a differentiable PDF which can be evaluated efficiently.
    \item It has a differentiable CDF which can be evaluated efficiently.
    \item It is flexible enough to approximate a broad class of conditional time-to-event distributions $p(t|x)$ with support over $\mathbb{R}^+$.
\end{enumerate}
The first two properties enable efficient training
using maximum likelihood and using stochastic gradients.
Examples of the last property are models that do not make assumptions like lognormality or proportional hazards, or that can capture multiple modes.

\subsection{Mixture Density Networks}
Mixture Density Networks (MDNs)~\citep{bishop1994mixture} form the key part of Survival MDNs. For a given $x$, MDNs model the conditional distribution $p(y|x)$ by mapping $x$ through a neural network to produce the weights and parameters of a mixture model.
 Mixture density networks are flexible approximators; for any given $x$, with enough components, MDNs can approximate a broad class of conditional densities $p(y|x)$ as closely as desired~\citep{bishop1994mixture}.  

In this work, we use Gaussian mixtures~\citep{reynolds2000speaker, reynolds2009gaussian}. A discussion on different base distributions can be found in \cref{appsec:base}. Assume we have $K$ components with weights $\{w_i\}_{i=1}^K$, means $\{\mu_i\}_{i=1}^K$ and standard deviations $\{\sigma_i\}_{i=1}^K$
 such that $\sum_{i=1}^K w_i = 1$. The PDF of the Gaussian Mixture Density Network is given by
\[
p\left(y|\{w_i,\mu_i,\sigma_i \}_{i=1}^K\right) = \sum_{i=1}^K w_i \mathcal{N}\left(y|\mu_i, \sigma_i^2\right),
\]
where we denote $\mathcal{N}(y|\mu_i, \sigma_i^2)$ as the density of a Gaussian distributed random variable with mean $\mu_i$ and variance $\sigma_i^2$. 

In mixture density networks, we build the conditional distribution by mapping the covariates $x$ to parameters of the Gaussian Mixture Model through deep neural networks:
\[
\{w_i(x), \mu_i(x), \sigma_i(x)\}_{i=1}^K = f_\theta(x),
\]
where $f_\theta$ is a trainable neural network with parameters $\theta$.
\subsection{Survival Mixture Density Networks}
We propose Survival Mixture Density Networks (Survival MDNs) to satisfy the properties we want for a survival model. 

The sampling process for Survival MDN on a given input $x$ is 
\begin{enumerate}
    \item Calculate $\{w_i(x), \mu_i(x), \sigma_i(x)\}_{i=1}^K = f_\theta(x)$.
    \item Sample $y$ according to the PDF $\sum_{i=1}^K w_i \mathcal{N}\left(y|\mu_i, \sigma_i^2\right)$. To do so, first sample a component $i$ with probability equal to $w_i$ and then sample from $\mathcal{N}(\mu_i, \sigma_i^2)$.
    \item Map $y$ to the event time $t$ using $t = g(y) = \log(1 + \exp(y))$.
\end{enumerate}
The invertible \texttt{softplus} function $g(y) = \log(1 + \exp(y))$ maps the sample from the mixture density network to the positive domain. Another common choice to map the input from $\mathbb{R}$ to $\mathbb{R}^+$ is \texttt{exp}. We choose \texttt{softplus} over \texttt{exp} for the reason that \texttt{exp} may place high density on very large times. 

Next, we show that the PDF and CDF of the Survival MDN is easy to compute. By the change of variables, the Survival MDN PDF at time $t$ for
input $x$ is:
\[
p(t|x)= \Big|\frac{d g^{-1}(t)}{dt}\Big|\Big(\sum_{i=1}^K w_i(x) \mathcal{N}\left(g^{-1}(t)|\mu_i(x), \sigma_i^2(x)\right)\Big).
\]
For the simple choice of the \texttt{softplus}, the absolute value term does not depend on the parameters of neural network $f_\theta$ so this term does not contribute to gradients used for log-likelihood training. The Survival MDN CDF at time $t$ can be computed easily as well. Denote $F(\cdot|\mu_i, \sigma_i^2)$ as the CDF of the $i$-th component in the Gaussian mixture model. Denote $F(t|x)$ as the CDF of the Survival MDN and $F_{\text{MDN}}(y|x)$ as the CDF of the underlying MDN. Since \texttt{softplus} is an increasing invertible function, we show that the CDF of the Survival MDN at time $t$ only requires evaluations
of the underlying Gaussian CDFs:

\begin{align*}
 F(t|x) &= F_{\text{MDN}}(g^{-1}(t)|x)  \\
 &= \int_{-\infty}^{g^{-1}(t)} \sum_{i=1}^K w_i(x) N\left(y|\mu_i(x), \sigma_i^2(x)\right) dy\\
 &= \sum_{i=1}^K w_i(x)  \int_{-\infty}^{g^{-1}(t)}N\left(y|\mu_i(x), \sigma_i^2(x)\right) dy   
\\
 &= \sum_{i=1}^K w_i(x) F\left(g^{-1}(t)|\mu_i(x), \sigma_i^2(x)\right)\\ 
\end{align*}
The evaluation of Gaussian CDFs can be done efficiently through the error function $\texttt{erf}(\cdot)$ which is the CDF of the standard normal distribution:
\[
F\left(g^{-1}(t)|\mu_i(x), \sigma_i^2(x)\right) = \texttt{erf}\left(\left(g^{-1}(t) - \mu_i(x)\right)/\sigma_i(x)\right).
\]
The \texttt{erf} function can be computed efficiently via common approximations~\citep{abramowitz1988handbook} and the Gaussian CDF is implemented in most packages. Now we have satisfied the first two desired properties (PDF and CDF). The last property, flexibility, follows since the Survival MDN maps time-to-event densities to densities over the reals via $y =g^{-1}(t)$ and a mixture density network with enough components and a wide and deep enough network can approximate a broad class of smooth densities $p(y|x)$ as closely as desired~\citep{bishop1994mixture}. For tabular data, the network in MDN is a feedforward neural network. Other types of networks can also be used. For example, for image data, one can use convolutional neural networks and for text data one can use transformers. Instead of logits for classification, these models produce the parameters of the Gaussian Mixture at the last layer in MDNs. 

\section{Simulation Study}
In this simulation experiment, we test Survival MDN and SODEN on a dataset where the proportional hazard assumption does not hold. We follow the simple simulation setting in SODEN \citep{tang2020soden}. There are two group of $x$'s, $x=0$ and $x=1$, and the ground truth survival function is:
\[
S(t|x) = \exp(-2t)\cdot I\{x=0\}  + \exp(-2t^2) \cdot I\{x=1\},
\]
where $I$ is the indicator function. The survival curves of the two groups cross so this survival distribution does not obey the proportional hazard (PH) assumption. Therefore, models that require the PH assumption cannot fit this dataset well. We generate $x$ from a Bernoulli distribution with probability 0.5 and then generate $t$ using the inverse CDF method. We sample the censored time uniformly on $[0,2]$. Instead of simulating a fixed dataset, we use an ``online'' training method; in each iteration, we generate a new set of 1024 datapoints. We use the likelihood function for training. We train for 10,000 iterations for both SODEN and Survival MDN. 

\begin{figure}
    \centering
    \includegraphics[width=\textwidth]{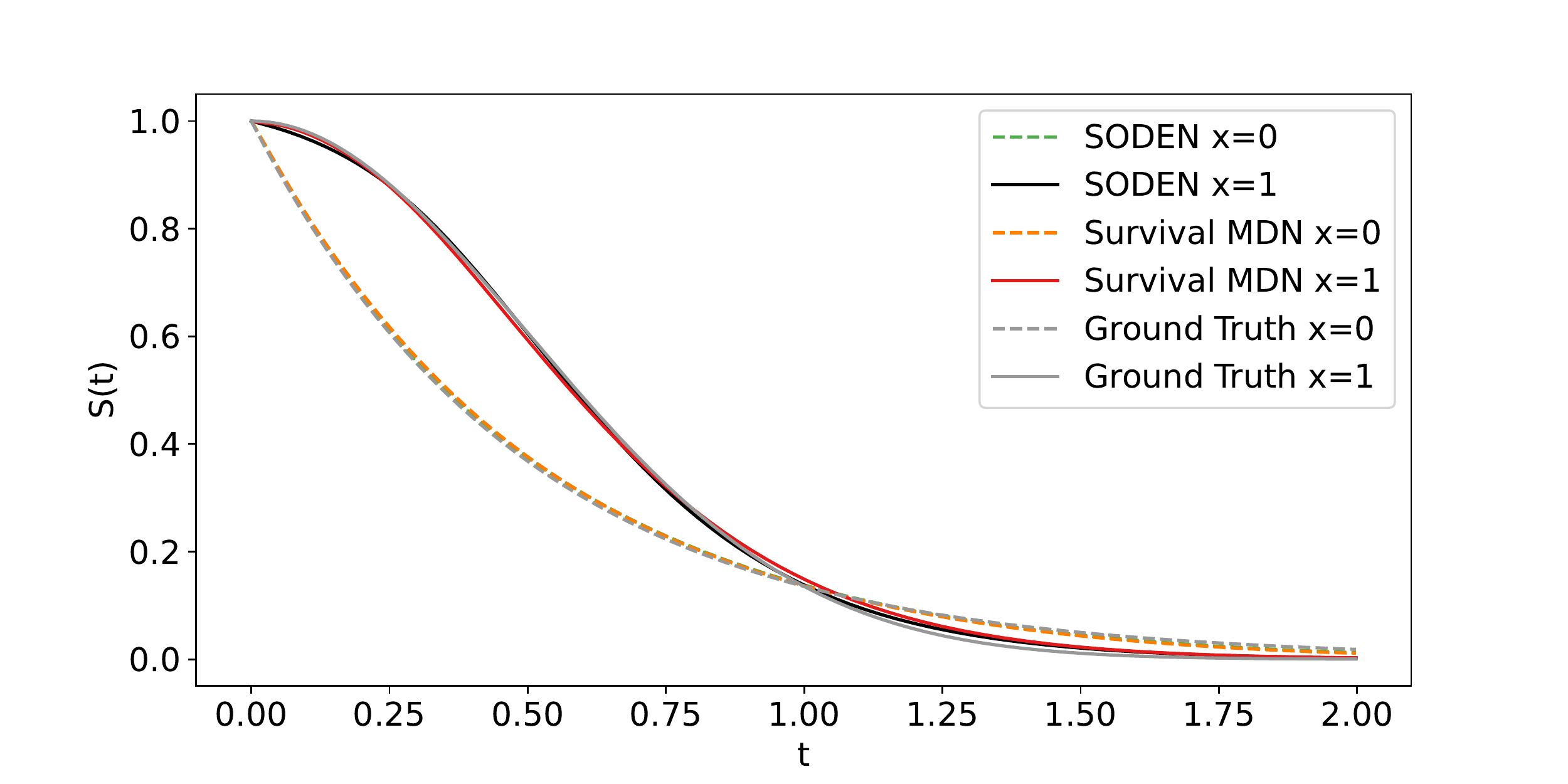}
    \caption{The survival functions of two groups. We show the survival function of two trained model SODEN and Survival MDN together with the ground truth on $x=0$ and $x=1$ separately. For $x=0$, three survival functions are so close to each other that the green curve for SODEN is covered by the blue curve for Survival MDN and the gray curve for the ground truth.}
    \label{fig:survival_func}
\end{figure}

We show the resulting survival functions and ground truth in \cref{fig:survival_func}. Both Survival MDN and SODEN's survival functions are close to the ground truth at both $x=0$ and $x=1$.

\section{Real World Experiments}
In this section, we compare Survival MDN with baselines Cox, DeepSurv, Cox-Time, Nnet-survival, DeepHit and SODEN. We use four different datasets: SUPPORT, METABRIC, GBSG and MIMIC. We evaluate all models on three different metrics: concordance, integrated binomial log-likelihood and integrated Brier score.
\subsection{Datasets}
We choose four different datasets: SUPPORT, METABRIC, GBSG and MIMIC. SUPPORT, METABRIC and GBSG are commonly used datasets for survival analysis, which can be found in the \texttt{pycox} package. MIMIC is a dataset we preprocessed from MIMIC-iv~\citep{johnson2020mimic} in PhysioNet~\citep{goldberger2000physiobank}. We describe the details of the datasets here:
\begin{itemize}
    \item SUPPORT: the Study to Understand Prognoses
Preferences Outcomes and Risks of Treatment. It has 14 features. There are 8,873 datapoints, 32\% of which are censored. We use the train/valid/test splits from SODEN~\citep{tang2020soden}\footnote{\label{note1}Available at https://github.com/jiaqima/SODEN}.
\item METABRIC: the Molecular Taxonomy of
Breast Cancer International Consortium. It has 9 features. There are 1,904 datapoints, 42\% of which are censored. We use the train/valid/test splits from the SODEN repository.
\item GBSG: The Rotterdam \& German Breast Cancer Study Group. It has 7 features. There are 2,232 datapoints, 43\% of which are censored.
\item MIMIC: The Medical Information Mart for Intensive Care. The SODEN repository does not provide the data files for MIMIC. We choose patients that are alive 24 hours after admission to ICU. We define the event as mortality after admission. We define the censored time as the ICU discharged time. We collect time series features within the 24-hour window after the admission together with static features. For time series features, we use the minimum, mean and maximum within the window. We remove the features that are missing for more than half of the datapoints. Finally, we extract 65 features after preprocessing including common labs and vitals. There are 53,612 datapoints, 82\% of which are censored. The SQL code that preprocesses the data from MIMIC-iv is attached in \cref{sec:sql}.
\end{itemize}
\begin{table}[h]
    \centering
    \begin{tabular}{cc|ccc}
    \toprule
       $P(C>\tau)$  & Model & $C_{\tau}^{td}(\uparrow)$ & IBLL$_{\tau}$($\uparrow$) & IBS$_{\tau}$($\downarrow$) \\ \midrule
       $10^{-8}$  & Cox &0.596 $\pm$ .002 & -0.568 $\pm$ .001& 0.194 $\pm$ .001 \\
                  & DeepSurv &0.609 $\pm$ .003& \textbf{-0.559} $\pm$ .002 & \textbf{0.190} $\pm$ .001\\
                  & Cox-Time & 0.607 $\pm$ .004 & 0.565 $\pm$ .002 & 0.191 $\pm$ .001 \\
                  & Nnet-Survival &  0.624 $\pm$ .003 & -0.570 $\pm$ .004 &0.193 $\pm$ .001  \\
                  & DeepHit & \textbf{0.631} $\pm$ .003 & -0.583 $\pm$ .006 & 0.197 $\pm$ .001  \\
                  & SODEN & 0.627 $\pm$ .003 & -0.563 $\pm$ .002 & 0.191 $\pm$ .001\\
                  & Survival MDN & 0.628 $\pm$ .003  & \textbf{-0.559} $\pm$ .002 &  \textbf{0.190} $\pm$ .002 \\  \midrule
       $0.2$  & Cox &0.596 $\pm$ .002 & -0.585 $\pm$ .001 & 0.201 $\pm$ .000\\
                  & DeepSurv & 0.609 $\pm$ .003 & -0.577 $\pm$ .002 &0.197 $\pm$ .001\\
                  & Cox-Time & 0.606 $\pm$ .004 & -0.583 $\pm$ .002 &0.199 $\pm$ .001 \\
                  & Nnet-Survival & 0.623 $\pm$ .003 & -0.586 $\pm$ .003 & 0.201 $\pm$ .001  \\
                  & DeepHit &  \textbf{0.630} $\pm$ .003 & -0.601 $\pm$ .006 & 0.205 $\pm$ .002 \\
                  & SODEN & \textbf{0.630} $\pm$ .003 & -0.601 $\pm$ .006 & 0.205 $\pm$ .002   \\
                  & Survival MDN & 0.628 $\pm$ .003  & \textbf{-0.575}$\pm$ .002  & \textbf{0.196} $\pm$ .001   \\ \midrule
       $0.4$  & Cox & 0.595 $\pm$ .002 & -0.602 $\pm$ .001 & 0.208 $\pm$ .001  \\
                  & DeepSurv &  0.608 $\pm$ .002 & -0.595 $\pm$ .002 &0.205 $\pm$ .001 \\
                  & Cox-Time & 0.605 $\pm$ .004 & -0.601 $\pm$ .002 & 0.207 $\pm$ .001  \\
                  & Nnet-Survival & 0.623 $\pm$ .003 & -0.602 $\pm$ .003 & 0.208 $\pm$ .001  \\
                  & DeepHit & \textbf{0.630} $\pm$ .003 & -0.619 $\pm$ .007 & 0.212 $\pm$ .002 \\
                  & SODEN &  0.626 $\pm$ .003 & -0.597 $\pm$ .002 & 0.205 $\pm$ .001\\
                  & Survival MDN & 0.628 $\pm$ .003 & \textbf{-0.593} $\pm$ .001  &  \textbf{0.204} $\pm$ .001  \\ \bottomrule
    \end{tabular}
    \caption{Evaluation of all models on SUPPORT with  concordance ($C_{\tau}^{td})$, integrated binomial log-likelihood (IBLL$_{\tau}$) and integrated Brier score (IBS$_{\tau}$). The \textbf{bold} number indicates the best performance. We report mean $\pm$ standard error on all metrics.}
    \label{tab:support}
\end{table}
\subsection{Baselines}
We consider the following baseline models:
\begin{itemize}
    \item Cox~\citep{cox1972regression}: A linear model with the proportional hazards assumption.
    \item DeepSurv~\citep{katzman2018deepsurv}: A deep model with the linear function in Cox replaced by neural networks.
    \item Cox-Time~\citep{katzman2018deepsurv}: A continuous time model that allows the relative risk in Cox to depend on time.
    \item Nnet-Survival~\citep{gensheimer2019scalable}: A discrete time model that models the conditional hazard in each time interval.
    \item DeepHit~\citep{lee2018deephit}: A deep discrete time model that further adds a rank-based loss to the likelihood as the training objective.
    \item SODEN~\citep{tang2020soden}: An ODE-based continous time model.
\end{itemize}
For Cox, we use the implementation in the Python package \texttt{lifelines}. For DeepSurv, Cox-Time, Nnet-Survival and DeepHit, we use the implementations in the Python package \texttt{pycox}. For SODEN, we use the implementation from the SODEN repository.

\begin{table}[h]
    \centering
    \begin{tabular}{cc|ccc}
    \toprule
       $P(C>\tau)$  & Model & $C_{\tau}^{td}(\uparrow)$ & IBLL$_{\tau}$($\uparrow$) & IBS$_{\tau}$($\downarrow$) \\ \midrule
       $10^{-8}$  & Cox &0.644 $\pm$ .006 & -0.508 $\pm$ .009& 0.169 $\pm$ .002 \\
                  & DeepSurv &0.635 $\pm$ .007& -0.517 $\pm$ .011 & 0.171 $\pm$ .003\\
                  & Cox-Time & 0.648 $\pm$ .007 & -0.511 $\pm$ .009 & 0.172 $\pm$ .003 \\
                  & Nnet-Survival &  0.666 $\pm$ .005 & -0.510 $\pm$ .007 &0.171 $\pm$ .002  \\
                  & DeepHit & \textbf{0.674} $\pm$ .006 & -0.514 $\pm$ .004 & 0.174 $\pm$ .002  \\
                  & SODEN & 0.661 $\pm$ .005 & -0.498 $\pm$ .008 & 0.167 $\pm$ .003\\
                  & Survival MDN & 0.667 $\pm$ .004  & \textbf{-0.489} $\pm$ .005 &  \textbf{0.165} $\pm$ .002 \\  \midrule
       $0.2$  & Cox & 0.639 $\pm$ .006 & -0.521 $\pm$ .006 &0.176 $\pm$ .002\\
                  & DeepSurv & 0.635 $\pm$ .006 & -0.530 $\pm$ .005 &0.179 $\pm$ .002 \\
                  & Cox-Time & 0.647 $\pm$ .005 & -0.531 $\pm$ .007 &0.179 $\pm$ .002 \\
                  & Nnet-Survival & 0.662 $\pm$ .004 & -0.523 $\pm$ .003 & 0.177 $\pm$ .001  \\
                  & DeepHit &  \textbf{0.671} $\pm$ .004 & -0.533 $\pm$ .003 & 0.182 $\pm$ .001 \\
                  & SODEN & 0.659 $\pm$ .003 & -0.516 $\pm$ .006 & 0.174 $\pm$ .002   \\
                  & Survival MDN & 0.662 $\pm$ .004  & \textbf{-0.510} $\pm$ .003  & \textbf{0.172} $\pm$ .001   \\ \midrule
       $0.4$  & Cox & 0.637$\pm$ .006 & -0.521 $\pm$ .006 & 0.175 $\pm$ .002  \\
                  & DeepSurv &  0.635 $\pm$ .006 & -0.526 $\pm$ .005 &0.178 $\pm$ .002 \\
                  & Cox-Time & 0.644 $\pm$ .005 & -0.526 $\pm$ .006 & 0.178 $\pm$ .002  \\
                  & Nnet-Survival & 0.660 $\pm$ .003 & -0.519 $\pm$ .003 & 0.176 $\pm$ .001  \\
                  & DeepHit &\textbf{0.668} $\pm$ .003 & -0.528 $\pm$ .003 & 0.180 $\pm$ .001 \\
                  & SODEN &  0.658 $\pm$ .004 & -0.528 $\pm$ .003 & 0.180 $\pm$ .001\\
                  & Survival MDN & 0.660 $\pm$ .002 & \textbf{-0.508} $\pm$ .003  &  \textbf{0.172} $\pm$ .001  \\ \bottomrule
    \end{tabular}
    \caption{Evaluation of all models on METABRIC with  concordance ($C_{\tau}^{td})$, integrated binomial log-likelihood (IBLL$_{\tau}$) and integrated Brier score (IBS$_{\tau}$). We report truncated metrics for $\tau$'s satisfying $P(C>\tau) = 10^{-8},  0.2, 0.4$. The \textbf{bold} number indicates the best performance. We report mean $\pm$ standard error on all metrics.}
    \label{tab:metabric}
\end{table}
\subsection{Evaluation Metrics}
We use the same evaluation metrics as SODEN~\citep{tang2020soden}. They are concordance, integrated binomial log-likelihood and Brier score. The implementations can be found in the SODEN repository. We briefly describe the three metrics here and refer to \citet{tang2020soden} for more detailed descriptions.

\paragraph{Concordance} The concordance index is originally proposed by \citet{harrell1984regression}. It measures the probability that the relative order of the event time of two observations matches the predicted survival probabilities. \citet{antolini2005time} further relaxes the proportional hazard assumption in Harrell's concordance to create time dependent concordance. 
Building off the inverse-weighting method in \cite{cheng1995analysis},
\citet{uno2011c} 
introduces inverse probability weighted concordance to remove the dependence on the censoring distribution. They use the survival function of the censoring time $G(t) = P(C>t)$ as the weight and the Kaplan-Meier estimator for $G(t)$.  Under the completely random censoring assumption $C \indep (T,X)$, the inverse probability weighted estimator is consistent. This assumption is
routinely made for evaluation, e.g.
in  \cite{kvamme2019time, tang2020soden,han2021inverse}. Due to the limited number of observations, the estimator of the inverse weight $1/\hat{G}(t)$ may be very large for some large-enough $t$. So \citet{uno2011c} introduce a truncated version of the concordance estimator within a pre-specified time interval $[0,\tau]$:
\[
C_{\tau}^{td} = \frac{\sum_{i:\Delta_i=1, u_i < \tau} \sum_{j, u_i < u_j} I\left(\hat{S}(u_i|x_i) < \hat{S}(u_i|x_j)\right)/\hat{G}^2(u_i)}{\sum_{i:\Delta_i=1, u_i <\tau} \sum_{j:u_i<u_j}1/\hat{G}^2(u_i)},
\]
where $I(\cdot)$ is the indicator function. Here $\tau$ is used to truncate the large times that have very small $\hat{G}(t)$. We choose three $\tau$'s that satisfy $\hat{G}(\tau) = 10^{-8}, 0.2, 0.4$. When $\hat{G}(\tau)=10^{-8}$, the truncated concordance is almost equal to the non-truncated version.

\paragraph{Integrated Brier Score} The Brier score (BS) measures the mean square error between the ground-truth label and the predicted probability for a binary classifier. It measures both the calibration and discriminative performance \citep{degroot1983comparison}. In survival analysis, we evaluate the Brier score at a given time $t$. The label is whether the patient survives after time $t$ and the predicted probability is the survival function. We also consider an inverse probability weighted estimator~\citep{graf1999assessment,gerds2006consistent} for the Brier score at time $t$:
\[
\text{BS}(t) = \frac{1}{N}\sum_{i=1}^N \left\{\frac{\hat{S}^2(t|x_i) I(u_i \leq t , \Delta_i=1)}{\hat{G}(u_i)} + \frac{(1-\hat{S}(t|u_i))^2 I(u_i > t)}{\hat{G}(t)}\right\}.
\]

To consider all times, we use an integrated BS (IBS) over time interval $[0, \tau]$:
\[
\text{IBS}_{\tau} = \frac{1}{\tau}\int_0^{\tau} \text{BS}(t) dt.
\]
To avoid extreme inverse weights, we also report results for $\tau$'s that satisfy $\hat{G}(\tau) = 10^{-8},  0.2, 0.4$. When $\hat{G}(\tau) = 10^{-8}$, $\tau$ is almost equal to the maximum time in the data.
\begin{table}[h]
    \centering
    \begin{tabular}{cc|ccc}
    \toprule
       $P(C>\tau)$  & Model & $C_{\tau}^{td}(\uparrow)$ & IBLL$_{\tau}$($\uparrow$) & IBS$_{\tau}$($\downarrow$) \\ \midrule
       $10^{-8}$  & Cox &0.645 $\pm$ .009 & -0.523 $\pm$ .009& 0.177 $\pm$ .004 \\
                  & DeepSurv &0.663 $\pm$ .007& -0.509 $\pm$ .010 & \textbf{0.172} $\pm$ .004\\
                  & Cox-Time & 0.654 $\pm$ .007 & -0.521 $\pm$ .009 & 0.176 $\pm$ .003 \\
                  & Nnet-Survival &  0.661 $\pm$ .006 & -0.516 $\pm$ .008 &0.174 $\pm$ .005  \\
                  & DeepHit & 0.665 $\pm$ .008 & \textbf{-0.504} $\pm$ .017 & 0.176 $\pm$ .005  \\
                  & SODEN &  0.661 $\pm$ .012 & -0.514 $\pm$ .017 & 0.173 $\pm$ .004\\
                  & Survival MDN & \textbf{0.668} $\pm$ .007  & \textbf{-0.504} $\pm$ .006 &  \textbf{0.172} $\pm$ .003 \\  \midrule
       $0.2$  & Cox & 0.645 $\pm$ .009 & -0.519 $\pm$ .007 &0.176 $\pm$ .003\\
                  & DeepSurv & 0.663 $\pm$ .007 & -0.505 $\pm$ .008 &0.170 $\pm$ .002 \\
                  & Cox-Time & 0.654 $\pm$ .007 & -0.517 $\pm$ .006 &0.175 $\pm$ .002 \\
                  & Nnet-Survival & 0.661 $\pm$ .006 & -0.509 $\pm$ .006 & 0.170 $\pm$ .003  \\
                  & DeepHit &  0.665 $\pm$ .008 & -0.510 $\pm$ .008 & 0.172 $\pm$ .004 \\
                  & SODEN & 0.661 $\pm$ .012 & -0.510 $\pm$ .009 & 0.172 $\pm$ .004   \\
                  & Survival MDN & \textbf{0.668} $\pm$ .007  & \textbf{-0.501} $\pm$ .006 &  \textbf{0.168} $\pm$ .002   \\ \midrule
       $0.4$  & Cox & 0.645$\pm$ .009 & -0.519 $\pm$ .007 & 0.176 $\pm$ .003  \\
                  & DeepSurv &  0.663$\pm$ .007 & -0.505 $\pm$ .008 &0.170 $\pm$ .002 \\
                  & Cox-Time & 0.654 $\pm$ .007 & -0.517 $\pm$ .006 & 0.175 $\pm$ .002  \\
                  & Nnet-Survival & 0.661 $\pm$ .006 & -0.509 $\pm$ .007 & 0.170 $\pm$ .003  \\
                  & DeepHit & 0.665 $\pm$ .008 & -0.510 $\pm$ .008 & 0.172 $\pm$ .004 \\
                  & SODEN &  0.661 $\pm$ .012 & -0.510 $\pm$ .009 & 0.172 $\pm$ .004\\
                  & Survival MDN & \textbf{0.668} $\pm$ .007  & \textbf{-0.500} $\pm$ .006 &  \textbf{0.168} $\pm$ .002  \\ \bottomrule
    \end{tabular}
    \caption{Evaluation of all models on GBSG with  concordance ($C_{\tau}^{td})$, integrated binomial log-likelihood (IBLL$_{\tau}$) and integrated Brier score (IBS$_{\tau}$). We report truncated metrics for $\tau$'s satisfying $P(C>\tau) = 10^{-8},  0.2, 0.4$. The \textbf{bold} number indicates the best performance. We report mean $\pm$ standard error on all metrics.}
    \label{tab:gbsg}
\end{table}

\paragraph{Integrated Binomial Log-Likelihood} Another common metric for survival analysis is the integrated binomial log-likelihood (IBLL). Different from IBS, IBLL uses binomial (Bernoulli) log-likelihood at each time step $t$:
\[
\text{BLL}(t) = \frac{1}{N} \sum_{i=1}^N \left\{\frac{\log(1-\hat{S}(t|x_i)I(u_i\leq t, \Delta_i=1)}{\hat{G}(u_i)}+ \frac{\log(\hat{S}(t|x_i)I(u_i > t) }{\hat{G}(t)}\right\}.
\]

The IBLL is defined by:
\[
\text{IBLL}_{\tau} = \frac{1}{\tau}\int_0^{\tau} \text{BLL}(t) dt.
\]
 We also report results for $\tau$'s satisfying $\hat{G}(\tau) = 10^{-8} ,  0.2, 0.4$. 

\subsection{Experimental Setup}
We randomly split datasets into
training, validation, and testing sets. We use the validation set to choose the best epoch from training and hyperparameters and report the results on the test set. For SUPPORT/METABRIC/GBSG, we use 10 splits (8 for training, 1 for validation and 1 for test). For MIMIC, we use 5 splits (3 for training, 1 for validation, and 1 for test) since MIMIC is a larger dataset. We use random search to create 100 independent trials for different hyperparameters. We use the optimizer RMSProp~\citep{tieleman2012lecture}.

 For Survival MDN, following \citet{sudarshan2020deep}, we use a three-layer neural network that maps the features to a latent representation, and then from the latent representation we use three layers to output $w$'s, $\mu$'s, $\sigma$'s separately. We use a \texttt{softmax} layer to ensure that the sum of $w$'s equals one and use an \texttt{exp} function to ensure the standard deviations $\sigma$'s are positive. We vary the number of mixture components from 5 to 20. Different architectures can be used depending on the input type.
 
 For other models, we vary the number of layers. Other hyperparameters include the hidden sizes, learning rate, batch normalization, momentum, dropout, and batch size. For DeepHit and Nnet-Survival, we vary the number of time intervals in addition. 
 For other hyperparameters, we use the same tuning ranges as in \cite{tang2020soden}. We show the tuning ranges in \cref{sec:tune}.
 
\subsection{Results}
We report the results on the four datasets in \cref{tab:support} (SUPPORT), \cref{tab:metabric} (METABRIC),  \cref{tab:gbsg} (GBSG), and \cref{tab:mimic} (MIMIC). For SUPPORT and METABRIC, we use the exact same splits as the SODEN repository so we use their results for the baselines.

For concordance, DeepHit has the best concordance on SUPPORT and METABRIC while the continuous time model Survival MDN has the best concordances on GBSG and MIMIC. For IBLL and IBS, Survival MDN has the best performance across all datasets. The IBLL and IBS care more about the exact survival probability prediction at each time. The discrete time model DeepHit may not yield an accurate estimate of the survival probability for a particular time since it does not distinguish the times inside one time interval. For the discrete time models, it is also challenging to choose the bin boundaries~\citep{kvamme2019continuous, tang2020soden, craig2021survival}. The discrete models' concordance on MIMIC is worse than that of SODEN and Survival MDN. Continuous time models Survival MDN and SODEN have similar performance on concordance on the four datasets since they are both flexible continuous time models. There is little difference among $\hat{G}(\tau) = 10^{-8}, 0.2, 0.4$ for the concordance, IBLL and IBS on small datasets SUPPORT, METABRIC and GBSG, which is the same observation in SODEN~\citep{tang2020soden}.

\begin{figure}[h]
    \centering
    \includegraphics[width=\textwidth]{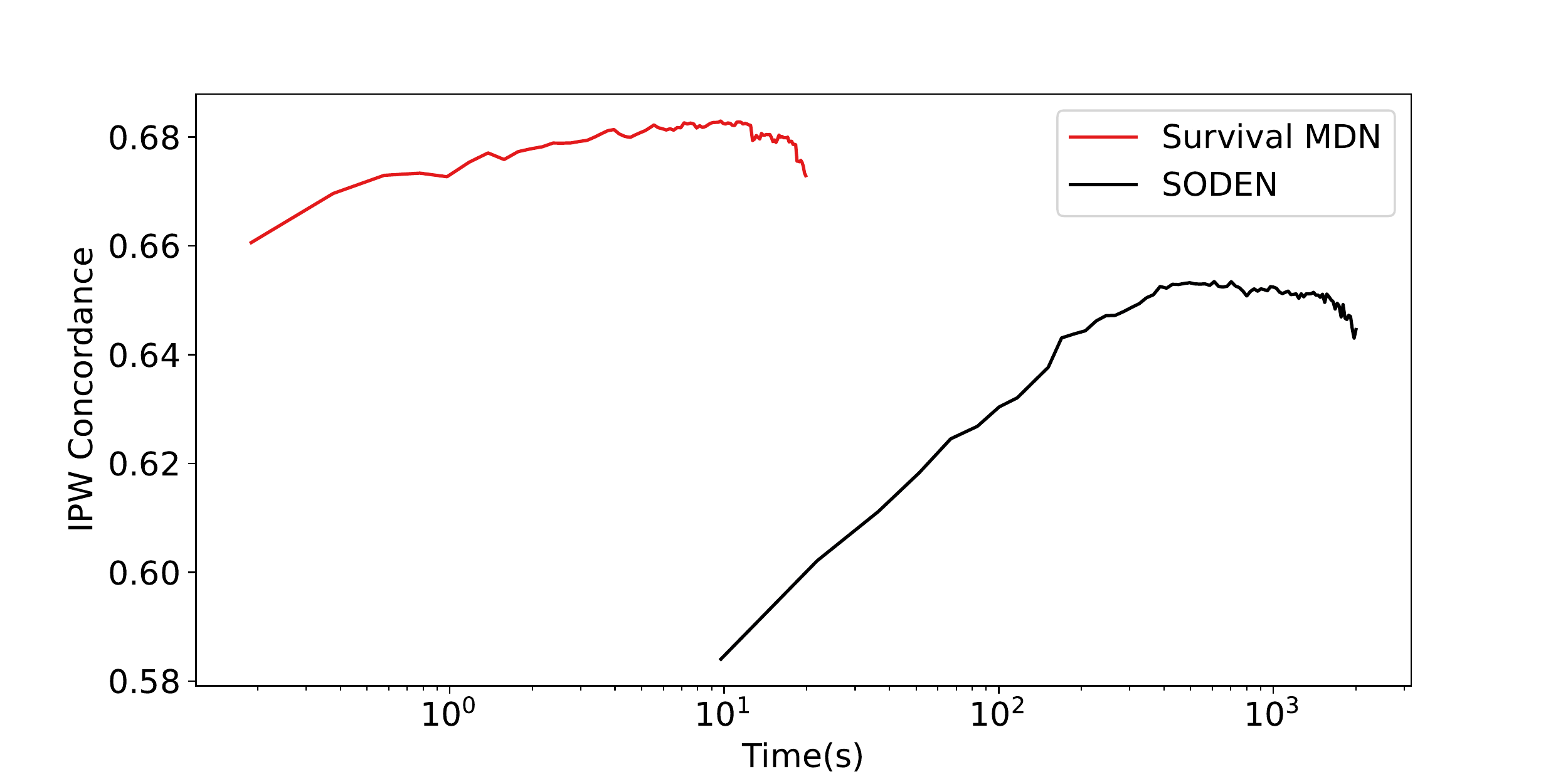}
    \caption{Comparison between Survival MDN and SODEN on IPW Concordance versus training time. The time is shown in log scale.}
    \label{fig:time_plot}
\end{figure}

\begin{table}[h]
    \centering
    \begin{tabular}{cc|ccc}
    \toprule
       $P(C>\tau)$  & Model & $C_{\tau}^{td}(\uparrow)$ & IBLL$_{\tau}$($\uparrow$) & IBS$_{\tau}$($\downarrow$) \\ \midrule
       $10^{-8}$  & Cox &0.642 $\pm$ .002 & -0.211 $\pm$ .001& 0.061 $\pm$ .001 \\
                  & DeepSurv &\textbf{0.663} $\pm$ .001& -0.212 $\pm$ .003 & 0.061 $\pm$ .001\\
                  & Cox-Time & 0.653 $\pm$ .001 & -0.210 $\pm$ .003 & 0.061 $\pm$ .001 \\
                  & Nnet-Survival &  0.649 $\pm$ .002 & -0.206 $\pm$ .000 &0.061 $\pm$ .001  \\
                  & DeepHit & 0.647 $\pm$ .002 & -0.206 $\pm$ .001 & 0.061 $\pm$ .001  \\
                  & SODEN & 0.659 $\pm$ .002 & \textbf{-0.204} $\pm$ .002 & 0.060 $\pm$ .001\\
                  & Survival MDN & 0.660 $\pm$ .002  & \textbf{-0.204} $\pm$ .002 &  \textbf{0.059} $\pm$ .001 \\  \midrule
       $0.2$  & Cox & 0.711 $\pm$ .004 & -0.473 $\pm$ .133 &0.091 $\pm$ .014\\
                  & DeepSurv & 0.734 $\pm$ .003 & -0.462 $\pm$ .150 &0.089 $\pm$ .015 \\
                  & Cox-Time & 0.726 $\pm$ .002 & -0.443 $\pm$ .126 &0.061 $\pm$ .001 \\
                  & Nnet-Survival & 0.722 $\pm$ .004 & -0.229 $\pm$ .004 & 0.066 $\pm$ .001  \\
                  & DeepHit &  0.719 $\pm$ .004 & -0.233 $\pm$ .004 & 0.066 $\pm$ .001 \\
                  & SODEN & 0.733 $\pm$ .002 & -0.229 $\pm$ .004 & \textbf{0.065} $\pm$ .001   \\
                  & Survival MDN & \textbf{0.736} $\pm$ .003  & \textbf{-0.228} $\pm$ .004  & \textbf{0.065} $\pm$ .001   \\ \midrule
       $0.4$  & Cox & 0.780 $\pm$ .002 & -0.588 $\pm$ .136 & 0.071 $\pm$ .031  \\
                  & DeepSurv &  0.797 $\pm$ .001 & -0.423 $\pm$ .202 &0.045 $\pm$ .018 \\
                  & Cox-Time & 0.790 $\pm$ .002 & -0.501 $\pm$ .267 & 0.037 $\pm$ .010  \\
                  & Nnet-Survival & 0.784 $\pm$ .003 & -0.082 $\pm$ .003 & \textbf{0.018} $\pm$ .001  \\
                  & DeepHit & 0.787 $\pm$ .003 & -0.083 $\pm$ .002 & 0.019 $\pm$ .001 \\
                  & SODEN &  \textbf{0.805} $\pm$ .005 & -0.084 $\pm$ .002 & 0.019 $\pm$ .001\\
                  & Survival MDN & \textbf{0.805} $\pm$ .001 & \textbf{-0.078} $\pm$ .002 &  \textbf{0.018} $\pm$ .001  \\ \bottomrule
    \end{tabular}
    \caption{Evaluation of all models on MIMIC with  concordance ($C_{\tau}^{td})$, integrated binomial log-likelihood (IBLL$_{\tau}$) and integrated Brier score (IBS$_{\tau}$). We report truncated metrics for $\tau$'s satisfying $P(C>\tau) = 10^{-8},  0.2, 0.4$. The \textbf{bold} number indicates the best performance. We report mean $\pm$ standard error on all metrics.}
    \label{tab:mimic}
\end{table}
The training time of SODEN is much longer than Survival MDN. We collect the training time of two models with the same hidden size 32 and number of layers 4 on METABRIC. We use the maximum number of components in the tuning range 20 for Survival MDN. We show the test concordance versus the training time for Survival MDN and SODEN on \texttt{GeForce RTX 2080 Ti} in \cref{fig:time_plot}. We can see that Survival MDN reached the peak of the test concordance much faster than SODEN. On average, each epoch of Survival MDN costs 0.20 seconds while each epoch of SODEN costs 23.82 seconds. Training Survival MDN is more than 100 time faster than SODEN.

\section{Discussion} 
Survival modeling plays an important role in risk estimation and clinical decision making. We propose Survival MDN, a simple flexible continuous time survival model. We combine two simple yet elegant tools---mixture densities and change of variables---to produce flexible survival models. While recent approaches achieve similar flexibility, it is achieved at the expense of training time, complexity, and inconvenient hyper-parameters. Without introducing such complexity, Survival MDNs achieve similar or better performance.

\paragraph{Limitations} Currently, the proposed model, Survival MDN, mainly considers Gaussian Mixtures. Though Gaussian Mixtures have universal approximation power, a combination of different base distributions, e.g. generalized logistics, in mixture density networks may improve performance. 
Regarding experimental evaluation, the marginal censoring assumption used in the reweighting estimators is common practice in the literature, but may not be appropriate.
Evaluation with censored data is impossible without assumptions, but it could be possible to improve evaluation by making conditional censoring assumptions.

\acks{
This work was made possible by the following grants/awards:
\begin{itemize}
    \item NIH/NHLBI Award R01HL148248 
    \item NSF Award 1922658 NRT-HDR: FUTURE Foundations, Translation, and Responsibility for Data Science.
    \item NSF CAREER Award 2145542
\end{itemize}
The authors thank Weijing Tang, Jiaqi Ma, Qiaozhu Mei and Ji Zhu for providing a great codebase. The authors thank Weijing Tang for a detailed explanation of the codebase.}

\bibliography{references}

\appendix
\section{\label{sec:sql}MIMIC SQL code}
\begin{verbatim}
select
-- ids
pat.subject_id as subject_id, adm.hadm_id as hadm_id,icu.stay_id as stay_id,
-- demographics
CASE WHEN pat.gender="M" THEN 1 ELSE 0 END as is_male,
CASE WHEN adm.ethnicity="WHITE" THEN 1 ELSE 0 END as is_white,
icu_detail.admission_age as age,
-- weight height 
fdw.weight ,
fdh.height ,
-- LOS
icu.los as los_icu_days,
icu_detail.los_hospital as los_hosp_days,
-- death
--icu_detail.icu_intime as icu_intime,
--icu_detail.dod as dod,
TIMESTAMP_DIFF(icu_detail.dod, icu_detail.icu_intime, HOUR) / 24 as time_to_death,
case 
    when icu_detail.dod is null then 0 
    else 1
end 
as death, 
-- vitals labs min max mean
vitals.*,
labs.*,
sofa.*
from `physionet-data.mimic_core.patients` pat   
inner join 
    `physionet-data.mimic_core.admissions` adm 
        on pat.subject_id=adm.subject_id
inner join 
    `physionet-data.mimic_icu.icustays` icu 
      on adm.subject_id=icu.subject_id
      and   
      adm.hadm_id=icu.hadm_id  
inner join 
     `physionet-data.mimic_derived.first_day_height` fdh
        on 
        adm.subject_id = fdh.subject_id  and icu.stay_id = fdh.stay_id 
inner join 
     `physionet-data.mimic_derived.first_day_weight` fdw
        on 
        adm.subject_id = fdw.subject_id  and icu.stay_id = fdw.stay_id 
inner join  
     `physionet-data.mimic_derived.icustay_detail` icu_detail   
        on  
        adm.subject_id=icu_detail.subject_id
        and 
        adm.hadm_id=icu_detail.hadm_id 
        and 
        icu.stay_id=icu_detail.stay_id 
inner join 
    `physionet-data.mimic_derived.first_day_sofa` sofa
    on 
        adm.subject_id=sofa.subject_id
        and 
        adm.hadm_id=sofa.hadm_id 
        and 
        icu.stay_id=sofa.stay_id 

inner join  
     `physionet-data.mimic_derived.first_day_vitalsign` vitals
     on 
    adm.subject_id=vitals.subject_id
    and 
    icu.stay_id=vitals.stay_id
inner join 
 `physionet-data.mimic_derived.first_day_lab` labs
     on 
    adm.subject_id=labs.subject_id
    and 
    icu.stay_id=labs.stay_id
where icu_detail.los_icu > 1
    and pat.gender is not null 
    and adm.ethnicity is not null
    and adm.ethnicity != "UNABLE TO OBTAIN"
    and adm.ethnicity != "UNKNOWN"    
\end{verbatim}
\section{\label{sec:tune}Tuning Ranges of Hyperparameters}
We show the search range of hyperparameters in 
\cref{tab:tune}.
\begin{table}[h]
    \centering
    \begin{tabular}{cc}
    \\\toprule
      Batch size   & $\{32, 64, 128, 256\}$ for METABRIC, GBSG  \\
      &  $\{128, 256, 512\}$ for SUPPORT \\
      &  $\{512, 1024\}$ for MIMIC \\
      Number of layers   & $\{1,2,4\}$ \\
      Hidden size & $[2^2, 2^7]$ \\
      Learning rate & $[10^{-4.5}, 10^{-1.5}]$ \\
      Weight decay & $[10^{-9}, 10^{-4}]$ \\
      Momentum & $[0.85, 0.99]$ \\
      Dropout & $\{0, 0.1, 0.5\}$ \\
      Batch normalization & $\{\text{True, False}\}$ \\
      $\alpha$ (Surrogate ranking loss in DeepHit) & $[0,1]$ \\
      $\sigma$ (Surrogate ranking loss in DeepHit) & $\{0.25, 1, 5\}$ \\
      Number of intervals& \{10, 50, 100, 200, 400\} for SUPPORT, METABRIC, GBSG \\
      (DeepHit, Nnet-survival)  & \{50, 100, 200, 400, 800\} for MIMIC \\
      \bottomrule
     \end{tabular}
    \caption{Tuning ranges of hyperparameters}
    \label{tab:tune}
\end{table}

\clearpage
 \section{\label{appsec:base}Discussion on Different Base Distributions}
 Here we compare Gaussian base with an altenative base, the generalized logistic distribution, on marginal data generations. We use the following form of the generalized logistic distribution: 
 \[
F(x; \alpha) = 1 - \frac{e^{-\alpha x}}{(1 + e^{-x})^{\alpha}}.
 \]
We also shift the generalized logistic distribution using scale and location. In this generalized logistic distribution, we have one more parameter $\alpha$ which can control the magnitude of the power. 

We consider three different marginal data generation cases:
 \begin{itemize}
     \item LogNormal distribution with $\mu = 0.1$ and $\sigma = 0.1$. LogNormal distribution is a common one researchers use in survival analysis. The variance is small in this data generation distribution.
    \item Student T distribution with degree of freedom one and transformed to positive values through softplus. Student T distribution has a heavy tail.
     \item Gamma distribution with shape 0.1 and scale 1. When shape is smaller than one, the Gamma distribution put a lot of mass on values close to zero. This may be hard for a mixture model to fit.
 \end{itemize}

 We sample the censored time uniformly from $[0,10]$. We still use an online training which generates a whole new batch data in every update step. 

\begin{figure}[h]
     \centering
     \includegraphics[width=8cm]{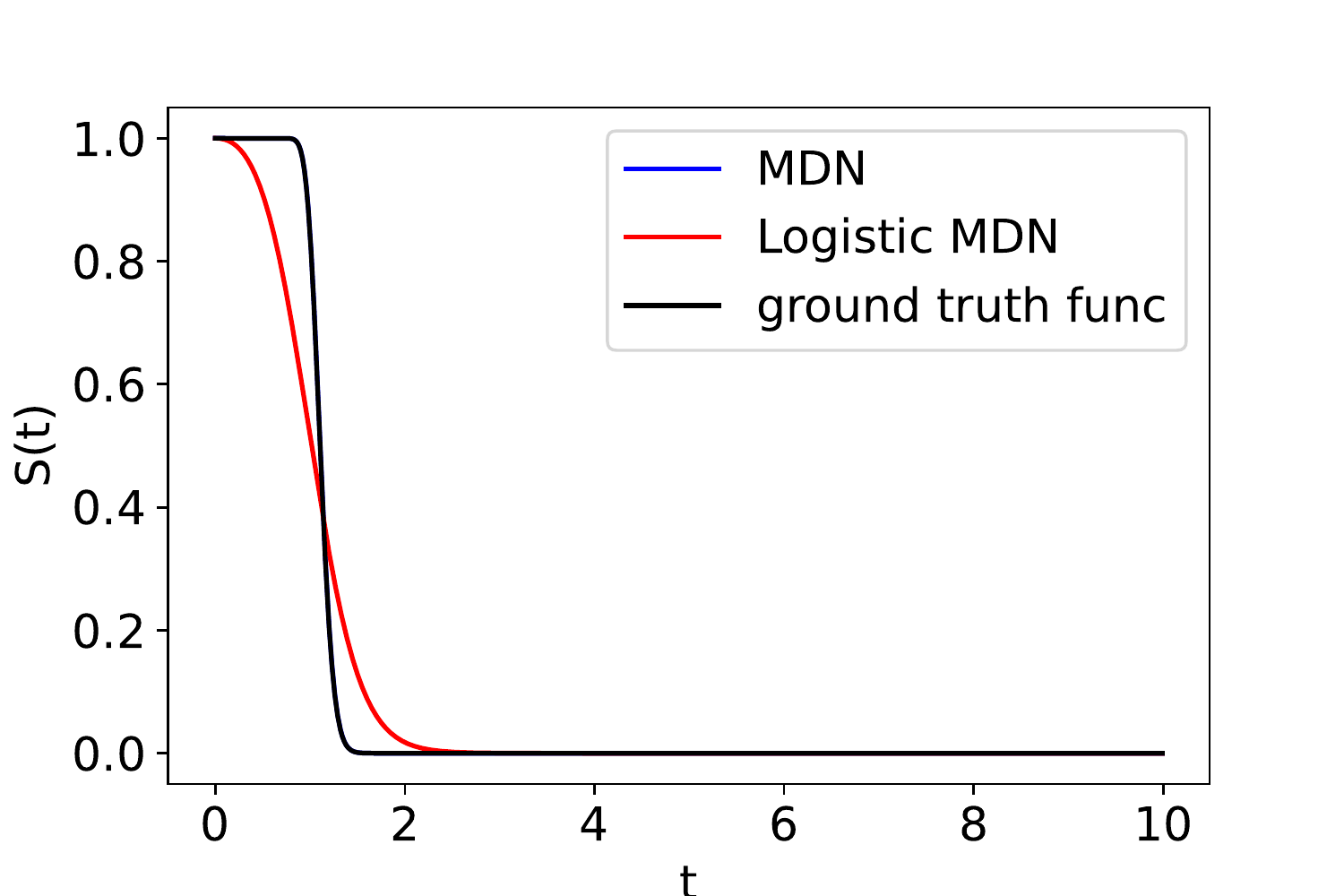}
     \caption{LogNormal Data.}
     \label{fig:lognormal}
 \end{figure}

 The results of fitting LogNormal data is shown in \cref{fig:lognormal}. The Gaussian base has survival functions overlapping with the ground truth but the generalized logistic base cannot fit it well. 
 
\begin{figure}[h]
     \centering
     \includegraphics[width=8cm]{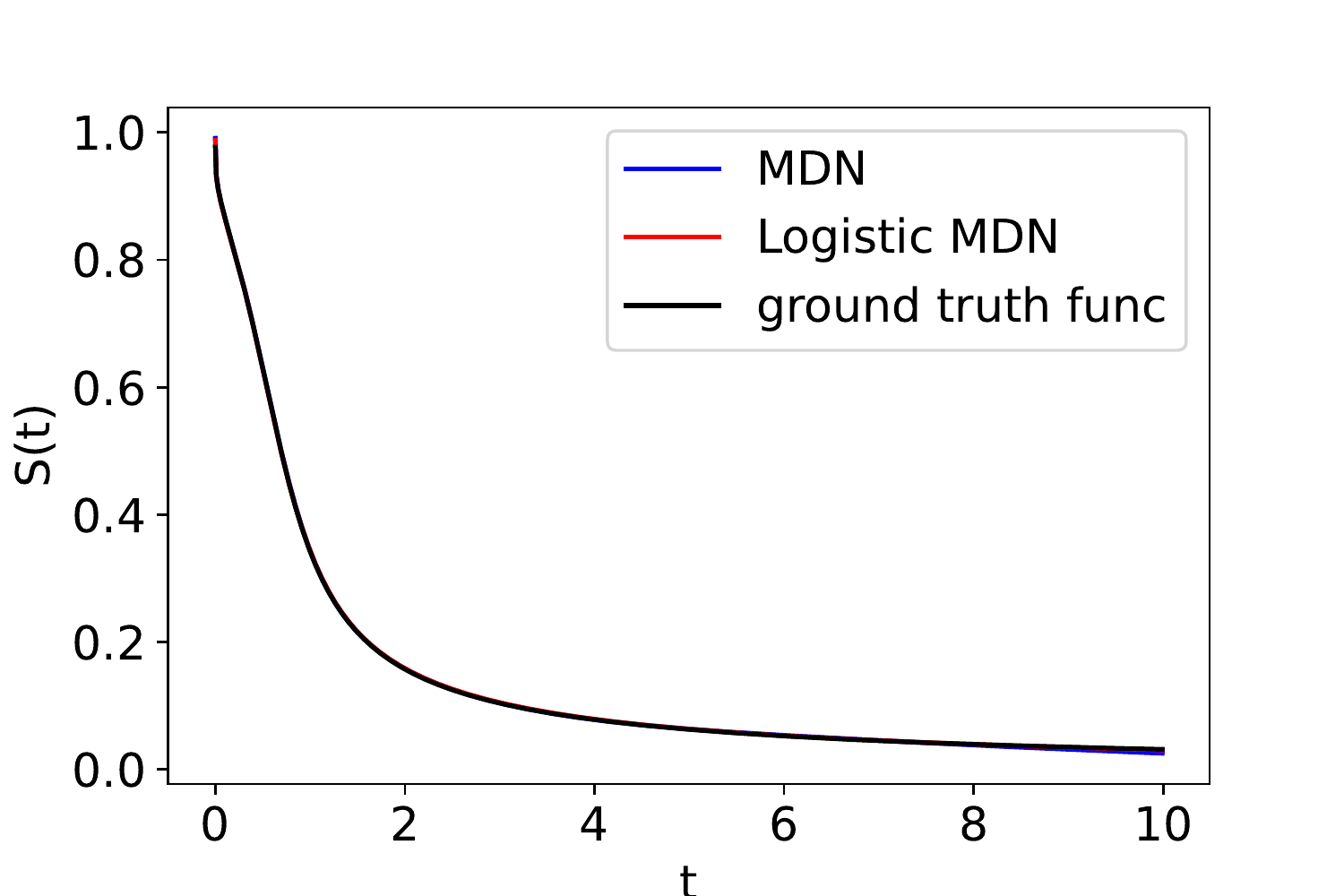}
     \caption{Student T + Softplus Data.}
     \label{fig:studentT}
 \end{figure}

 The result of fitting student T data is shown in \cref{fig:studentT}. For heavy tailed student T, both Gaussian base and generalized logistic base can also fit it well with survival functions overlapping the ground truth.
 
\begin{figure}[h]
     \centering
     \includegraphics[width=8cm]{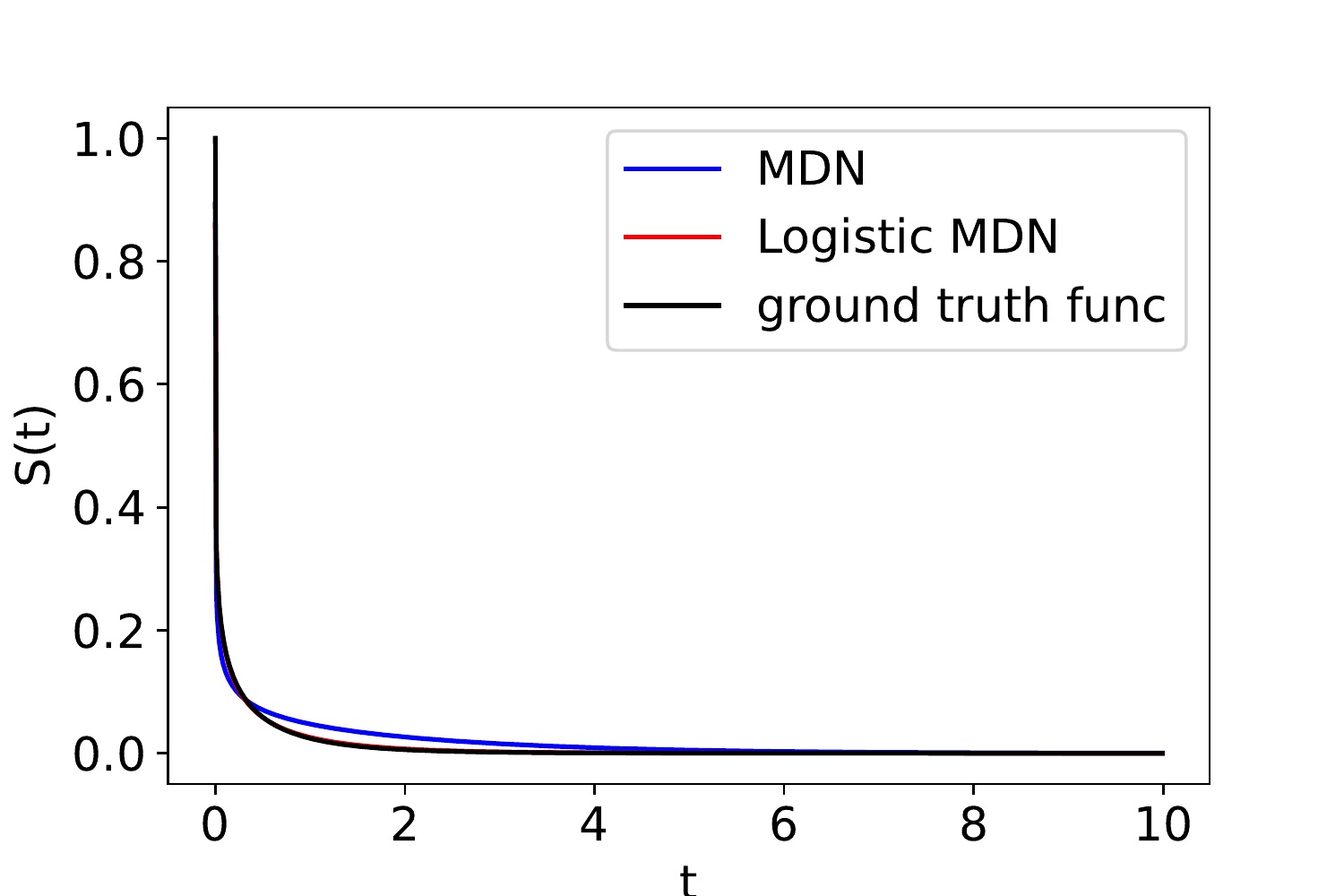}
     \caption{Gamma Data.}
     \label{fig:gamma}
 \end{figure}

  The result of fitting Gamma data is shown in \cref{fig:gamma}. The generalized logistic base can fit the Gamma data well while there is some gap between the ground truth and the Gaussian base survival function. In Gamma data with a small shape, the generalized logistic base is a better choice. 
\end{document}